\documentclass{article}
\usepackage{ijcai25}

\usepackage{times}
\usepackage{soul}
\usepackage{url}
\usepackage[hidelinks]{hyperref}
\usepackage[utf8]{inputenc}
\usepackage[small]{caption}
\usepackage{graphicx}
\usepackage{amsmath}
\usepackage{amsthm}
\usepackage{booktabs}
\usepackage{algorithm}
\usepackage{algorithmic}
\usepackage{amssymb,amsfonts}
\usepackage[switch]{lineno}
\usepackage{indentfirst}
\usepackage{subcaption}
\usepackage{forest}
\usetikzlibrary{arrows.meta,shapes,positioning,shadows,trees}

\tikzset{
    basic/.style  = {draw, text width=2cm, drop shadow, font=\sffamily, rectangle},
    root/.style   = {draw, thick, rounded corners=5pt, align=center, 
                     fill=blue!25, text width=0.7cm, font=\sffamily, 
                     minimum height=1.5cm, inner sep=2pt},
    onode/.style = {basic, thin, rounded corners=2pt, align=center, fill=green!10,text width=7em,},
    tnode/.style = {basic, thin, align=left, fill=pink!20, text width=15em, minimum height=0.6cm,rounded corners=0.15cm},
    xnode/.style = {basic, thin, rounded corners=2pt, align=center, fill=blue!20,text width=5cm,},
    wnode/.style = {basic, thin, align=left, fill=pink!10!blue!80!red!10, text width=10.5em},
    edge from parent/.style={draw=black, edge from parent fork right}
}

\title{Trustworthy GNNs with LLMs: A Systematic Review and Taxonomy}

\author{
Ruizhan Xue$^{1}$\thanks{Equal Contribution}
\and
Huimin Deng$^{1*}$\and
Fang He$^{1}$\\
Maojun Wang$^{1}$\and
Zeyu Zhang$^{1}$\thanks{Corresponding author (zhangzeyu@mail.hzau.edu.cn)}
\affiliations
$^1$National Key Laboratory of Crop Genetic Improvement\\
Hubei Hongshan Laboratory, Huazhong Agricultural University
}

\begin{document}

\maketitle

\begin{abstract}
    With the extensive application of Graph Neural Networks (GNNs) across various domains, their trustworthiness has emerged as a focal point of research. Some existing studies have shown that the integration of large language models (LLMs) can improve the semantic understanding and generation capabilities of GNNs, which in turn improves the trustworthiness of GNNs from various aspects. Our review introduces a taxonomy that offers researchers a clear framework for comprehending the principles and applications of different methods and helps clarify the connections and differences among various approaches. Then we systematically survey representative approaches along the four categories of our taxonomy. Through our taxonomy, researchers can understand the applicable scenarios, potential advantages, and limitations of each approach for the the trusted integration of GNNs with LLMs. Finally, we present some promising directions of work and future trends for the integration of LLMs and GNNs to improve model trustworthiness.
\end{abstract}

\section{Introduction}

\begin{figure}
    \centering
    \includegraphics[width=1.05\linewidth]{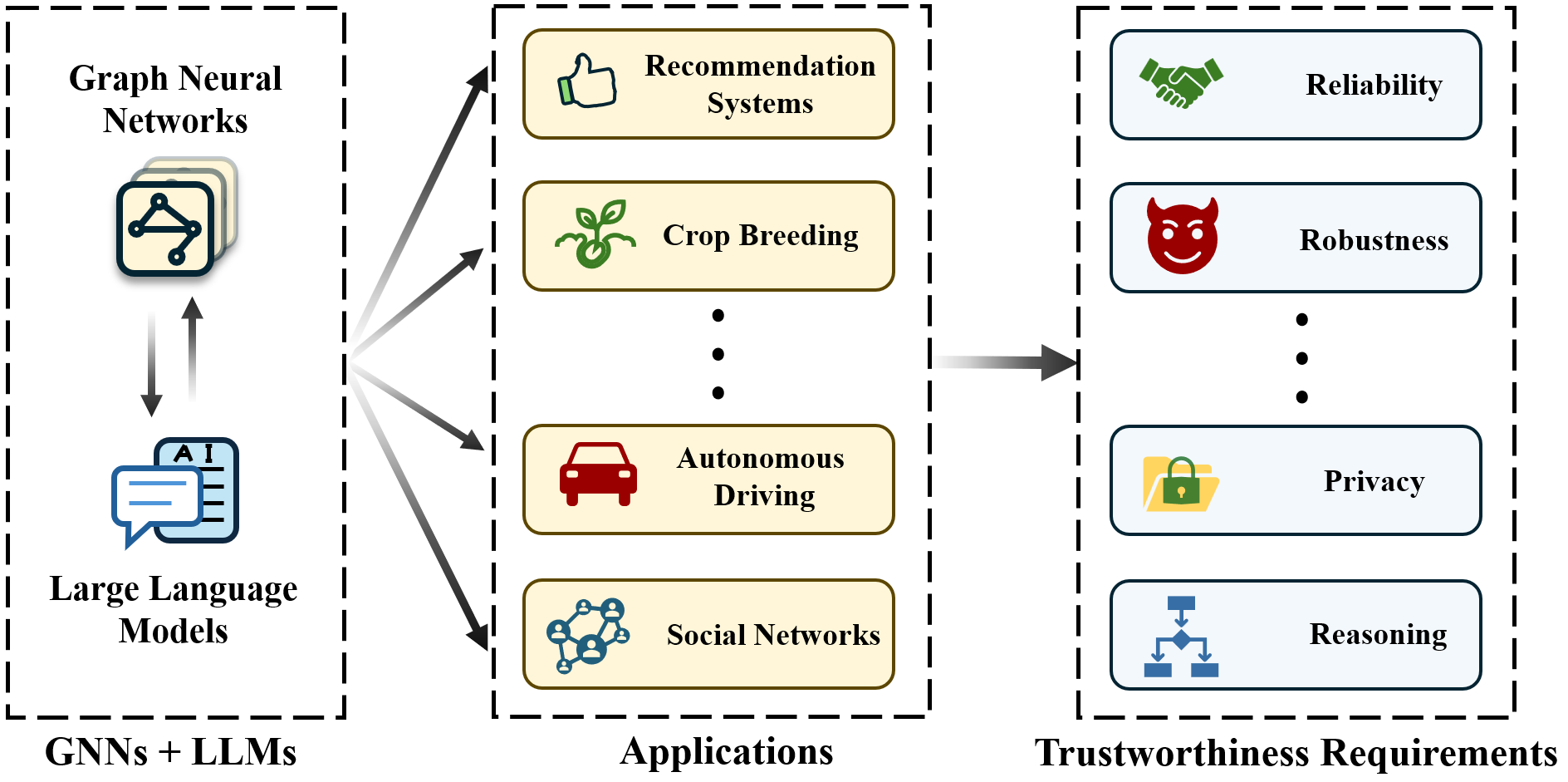}
    \caption{Applications of the integration of graphs and LLMs have driven the increased demand for model trustworthiness.}
    \label{fig:1}
\end{figure}

Graphs are data structures that are widely used in a variety of real-world scenarios \cite{xia2021graph}. Graph Neural Networks (GNNs) \cite{wu2020comprehensive} have achieved remarkable success in many fields due to their powerful modeling ability for graph-structured data, such as autonomous driving \cite{xiao2023review}, recommendation
systems \cite{zhang2023contrastive}, and crop breeding \cite{pan2024csgdn}
. With the deployment of GNNs in this highly sensitive fields, the trustworthiness of GNNs decisions has become a key bottleneck. GNNs are increasingly expected to be reliable, robust, and privacy-preserving to gain trust.

In recent years, significant progress has been made in large language models (LLMs) such as GPT \cite{brown2020language} and DeepSeek \cite{guo2025deepseek}. These variants have shown superior performance in many natural language processing tasks, such as sentiment analysis, machine translation, and text classification \cite{zhao2023survey}. Beyond traditional NLP applications, there is growing interest in using LLMs to process various data modalities, such as text-attributed graphs (TAGs) \cite{yang2021graphformers}.

Recent studies have shown that integrating LLMs into GNNs can substantially enhance node representations and improve model performance \cite{he2023harnessing,chen2024exploring}. This naturally raises an important question: Can the integration of LLMs and GNNs also enhance the trustworthiness of graph-based models? More specifically, how can LLMs and GNNs be effectively combined to improve model trustworthiness?

\textbf{LLMs help in trusted GNN-related tasks.} LLMs have significantly transformed how we interact with graph data, particularly in scenarios where nodes contain rich textual attributes. The integration of LLMs and graphs has been shown to be successful in a variety of graph-related tasks \cite{li2023survey}. As illustrated in Figure \ref{fig:1}, the growing adoption of this integration has led to increasing demands for trustworthy models. Numerous studies have shown that LLM-augmented GNNs can enhance trustworthiness. For example, LLM4RGNN \cite{zhang2024can} leverages the inferential capabilities of LLMs to identify malicious edges and recover missing important. 

\begin{figure}
    \flushleft
    \resizebox{0.48\textwidth}{!}{
    \begin{forest} for tree={
        grow=east,
        growth parent anchor=east,
        parent anchor=east,
        child anchor=west,
        edge path={\noexpand\path[\forestoption{edge}, >={latex}] 
         (!u.parent anchor) -- +(5pt,0pt) |- (.child anchor)
         \forestoption{edge label};}
    }
    [\rotatebox{90}{Trustworthy GNN with LLM}, root,anchor=center
        [Reasoning\\ (Explainability), onode
            [VGRL \cite{ji2024vgrl}, tnode]
            [LLMEP  \cite{zhang2024llmexplainer}, tnode]
            [GREASELM \cite{zhang2024greaselm}, tnode]
            [Graphllm \cite{chai2023graphllm}, tnode]
            [LLMRG \cite{wang2023enhancing}, tnode]
            ]
        [Robustness, onode
            [LLMGRobustness \cite{guo2024learning}, tnode]
            [LLM4RGNN \cite{zhang2024can}, tnode] ]
        [Privacy, onode
            [DLLM-GNN \cite{pan2024distilling}, tnode] ] 
        [Reliability, onode
            [LLMDGCN \cite{li2024llm}, tnode] 
            [GraphEdit \cite{guo2024graphedit}, tnode] ]   
    ]
    \end{forest}}
    \caption{A taxonomy of trustworthy GNNs with LLM.}
    \label{fig:2} 
\end{figure}
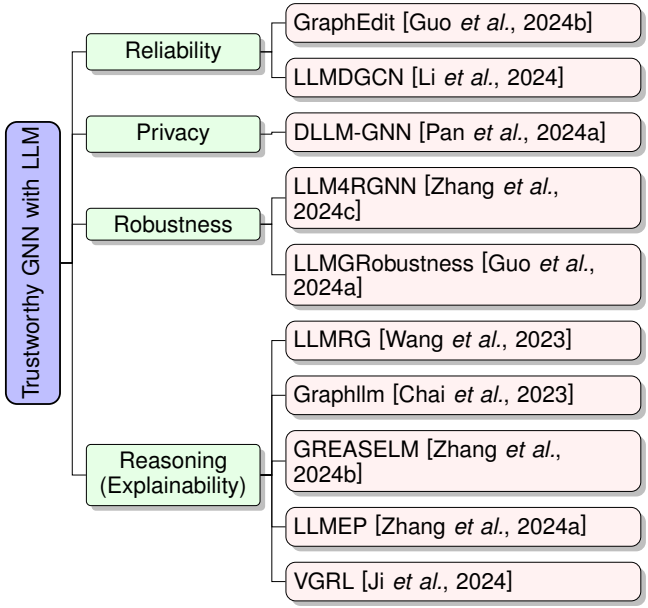

The integration of graphs and LLMS has a significant impact on the trustworthiness of graph-related tasks from different perspectives. To provide a systematic overview, as depicted in Figure \ref{fig:2}, we categorize existing works on enhancing trustworthiness through LLM-GNN integration into four key dimensions: reliability, robustness, privacy, and reasoning.


\textbf{Motivations.} With the growing applications of the integration of LLMs and GNNs, researchers are increasingly concerned about the trustworthiness of models. Although several approaches exist for integrating LLMs and GNNs to improve trustworthiness, these rapidly developing approaches often lack systematic organization and comparison. Presently, some papers have systematically explored the development of trustworthy GNNs \cite{zhang2022trustworthy}, but in the specific area of integrating LLMs and GNNs to improve model trustworthiness, no one has systematically investigated. In this survey, we aim to provide a timely survey summarizing these efforts focusing on ``Trustworthy GNNs with LLMs".

\textbf{Contributions.} The contributions of this survey can be summarized from the following three aspects. 
\begin{itemize}
    \item \textbf{A new taxonomy.} We are the first to propose a taxonomy in the specific area of the integration of LLMs and GNNs to improve model trustworthiness and classify these approaches. This taxonomy categorizes existing work into four categories.
    \item \textbf{A comprehensive review.} Based on the taxonomy, we provide a comprehensive overview of existing work on the integration of LLMs and GNNs to improve trustworthiness and point out their limitations.
    \item \textbf{Future Directions.} We propose promising research directions and future trends in this area.
\end{itemize}

\section{Preliminary}
In this section, we first introduce some notations throughout the paper. Then, we will further elaborate on three concepts that are closely related to this survey: Graph Neural Networks, Large Language Model, and Trustworthiness LLM-GNN.

\subsection{Notations}

We represent a graph as \( \mathcal{G} = (\mathcal{V}, \mathcal{E}) \), where \( \mathcal{V} = \{v_1, \ldots, v_N\} \) denotes the set of \( N \) nodes, and \( \mathcal{E} \subseteq \mathcal{V} \times \mathcal{V} \) represents the set of edges connecting these nodes. The graph can be either a plain graph (without attributes) or an attributed graph (with node features).

In the case of an attributed graph, each node \( v_i \) is associated with a feature vector \( x_i \in \mathbb{R}^d \), which represents the \(d \)-dimensional attribute of the node. All node attributes are collected as \( X = \{x_1,x_2 \ldots, x_N\} \).

The structure of the graph is captured by an adjacency matrix \( A \in \mathbb{R}^{N \times N} \), where each entry is defined as :

\[
A_{ij} =
\begin{cases} 
1, & \text{if nodes } v_i \text{ and } v_j \text{ are connected}, \\
0, & \text{otherwise}.
\end{cases}
\]

This adjacency matrix provides a mathematical representation of the graph's connectivity indicating which nodes share direct connections.

\subsection{Graph Neural Networks}

The core of Graph Neural Networks (GNNs) lies in updating node representations through the message-passing mechanism. GNNs utilize the message-passing mechanism to learn node representations that capture both node features and graph topology information. A GNN updates a node's representation by aggregating information from its neighboring nodes. Consequently, a k-layer GNN model captures local graph information within the k-hop neighborhood of the central node.
where:
\begin{itemize}
    \item \( h^{(k)}_v \) is the representation of node \( v \) at the \( k \)-th layer,
    \item \( \text{COMBINE}^{(k)} \) is the function to combine the node's own features with aggregated features from its neighbors,
    \item \( \text{AGGREGATE}^{(k-1)} \) aggregates the representations of neighbors \( \mathcal{N}(v) \) of node \( v \) at layer \( k-1 \).
\end{itemize}

Through message passing, GNNs can aggregate information from neighboring nodes. Additionally, by considering adjacency relationships when computing node representations, GNNs learn the entire graph's topology. Leveraging these properties, GNNs can perform various graph analysis tasks, including node classification, link prediction, graph classification, and community detection.

\textbf{Node classification} aims to assign a category label to each node in the graph based on its structure and node features. In this task, a subset of node labels (training set) is known and used for training, while the remaining node labels (test set) are predicted. For example, node classification can help determine whether a user is a``gamer" or a ``movie enthusiast".

\textbf{Link Prediction} seeks to determine whether a connection (edge) exists between two nodes or to estimate the weight or type of a relationship represented by the edge. This task is widely used to infer unknown or potential edges, such as predicting missing links in a drug knowledge graph.

\textbf{Graph Classification} assigns a category label to an entire graph based on its structural and attribute information. The goal is to learn a mapping from graph features to class labels, enabling classification across different graph instances.

\textbf{Community Detection} involves grouping nodes such that those within the same group are more closely related to each other than to nodes in other groups. The primary purpose is to partition the graph into meaningful substructures, such as identifying functional modules \cite{jiang2021graph} among genes or proteins.

\subsection{Large Language Model}

Two influential surveys \cite{zhao2023survey,motlagh2023cybersecurity} on distinguishing large language models (LLMs) from pre-trained language models (PLMs) differentiate them from two perspectives: model size and training approach. In terms of scale, LLMs are large language models with billions of parameters, such as GPT \cite{brown2020language} and Deepseek \cite{guo2025deepseek}, whereas PLMs are pre-trained models with millions of parameters. Notably, compared to PLM models, LLMs can learn new tasks from a small set of examples in the prompt during the inference stage without requiring additional fine-tuning.
Notably, compared to the emergent abilities of PLMs, LLMs exhibit new capabilities, including in-context learning, instruction following, and multi-step reasoning.

    In-context learning \cite{radford2021learning,dong2022survey}: During the inference stage, LLMs can learn new tasks from a small set of examples presented in the prompt without requiring additional fine-tuning.
    Instruction following: After undergoing instruction tuning, LLMs can understand and execute new task instructions without explicit examples.
    Multi-step reasoning: LLMs possess multi-step reasoning capabilities, allowing them to break down complex tasks into intermediate reasoning steps, as demonstrated in the Chain-of-Thought (CoT) \cite{wei2022chain} prompting.
This set of capabilities enables LLMs to handle complex tasks through few-shot or zero-shot learning.

The remarkable capabilities of LLMs are attributed to training on massive datasets and the design of the Transformer architecture with a large number of parameters. The Transformer architecture is the core technology of LLMs, with its multi-head self-attention mechanism and parallelized computation providing strong contextual understanding and the ability to capture long-range dependencies.

Another standout capability of LLMs is answer extraction from knowledge bases, allowing them to respond to specific questions about people, events, and places. They also exhibit the ability to perform logical reasoning based on multiple pieces of information, leveraging the extensive domain knowledge learned during training to draw relevant conclusions.

\subsection{Trustworthiness LLM-GNN}

The remarkable achievements of GNNs rely heavily on big data, as a large amount of sensitive data is collected from users to develop powerful GNN models for various services in critical domains. This raises potential privacy issues for GNN models. Membership inference attacks can determine whether certain users are included in the training data, threatening user privacy. Additionally, other privacy attack methods, such as link inference and attribute inference, can extract user information from pre-trained models.

In terms of trustworthy GNNs, several challenges persist in current research fields. For instance, due to the black-box nature of GNNs, their interpretability is insufficient. Moreover, the performance of trustworthy GNNs can deteriorate when dealing with perturbed data. LLMs can leverage their semantic capabilities to assist GNNs in providing rich sample interpretations in low-sample environments and can also help LLM-GNN models output readable reasoning processes to enhance interpretability.

\section{Proposed Taxonomy}
We propose a taxonomy, as shown in Figure \ref{fig:2}, that classifies existing representative techniques involving the integration of LLMs with GNNs to enhance model trustworthiness into four main categories: (1) Reliability, (2) Robustness, (3) Privacy, and (4) Reasoning. In the following sections, we provide a comprehensive survey based on these four categories.

\subsection{Reliability}
\begin{figure}
    \centering
    \includegraphics[width=1\linewidth]{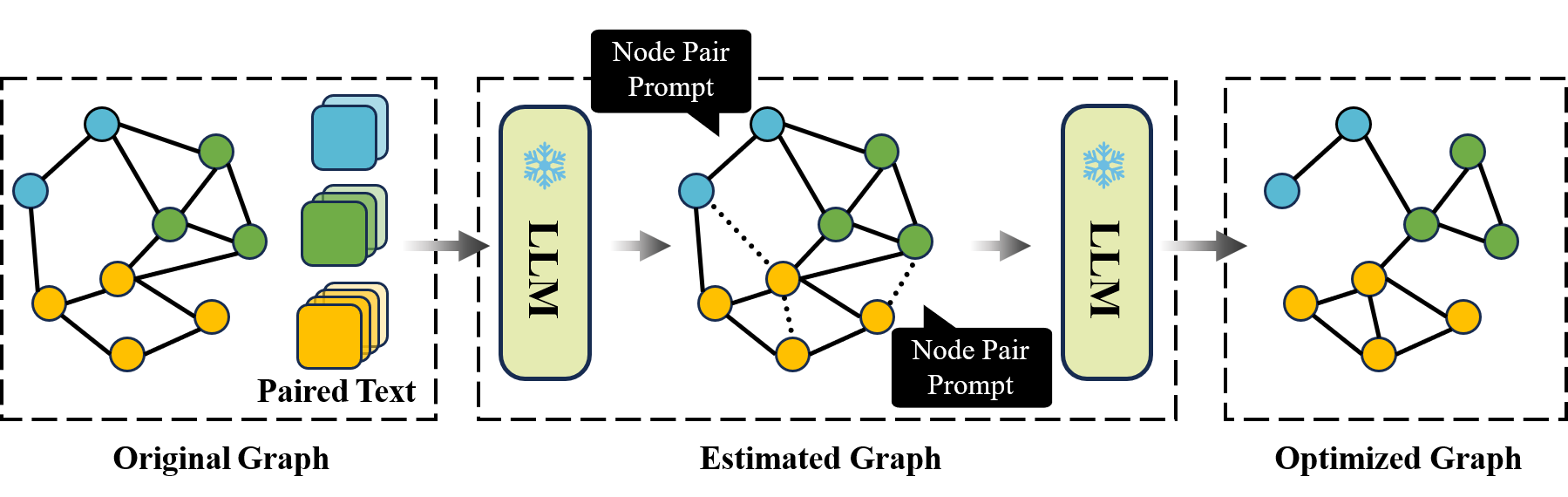}
    \caption{Train an LLM-based edge predictor to compute candidate edges; use a prompting approach to allow the LLM to evaluate the augmented adjacency matrix and determine the final edges.}
    \label{fig:3}
\end{figure}
The reliability of trustworthy GNNs lies in their ability to handle potential non-adversarial threats effectively, such as inherent noise and distribution shift. 

Inherent noise refers to unavoidable noise in graph data, which can be categorized into structural noise, attribute noise, and label noise. Real-world graph data often contain inherent noise due to errors in data measurement and collection. For instance, structural noise may be introduced in protein-protein interaction networks \cite{fionda2019networks}due to errors during data collection. In social networks, users may intentionally provide false information for privacy reasons, leading to attribute noise. Label noise is similar to data collection, as errors may also occur during the process of labeling nodes or graphs, especially when the labeling requires expertise in specific domains \cite{ji2022drugood}.

For structural noise, one feasible approach is to assign learnable weights to each node during node information aggregation in GNNs, allowing the model to focus on task-relevant connections. This can be achieved by adopting a self-attention mechanism, such as Graph Attention Networks (GAT) \cite{chen2020iterative}, to adjust the aggregation weights in GNNs.However, such methods cannot effectively integrate textual information into the denoising process of GNNs, whereas LLMs can generate high-quality pseudo labels, reducing the impact of label noise.

Another approach is to utilize neural networks to learn the subgraph distribution and sample task-relevant subgraphs from it, thereby removing irrelevant edges. This involves \cite{luo2021learning} using multi-layer neural networks to predict and prune irrelevant edges, extracting subgraphs from a learned distribution to enhance model performance.

Current applications of LLMs in trustworthy GNNs primarily focus on addressing inherent noise. LLMs can assist GNNs in mitigating structural noise in graph data. These applications mainly follow two directions: (1) LLMs are used to generate node embeddings, virtual nodes, and pseudo-labels, providing supplementary information for graph data. This approach emphasizes data generation and semantic augmentation. (2) LLMs directly interact with graph structures (in Figure \ref{fig:3}), such as adding or removing edges, thereby influencing graph structure learning. By understanding the context of the graph or specific task requirements, LLMs can generate suggestions for graph edits.

The first approach leverages the strong performance of LLMs in zero-shot and few-shot scenarios, enabling trustworthy GNNs to handle inherent noise better \cite{li2024llm}. Through the semantic capabilities of LLMs, missing labels in the graph can be supplemented, and textual descriptions can enrich node information. With the augmented information, GNNs can predict and complete edges between nodes. The second approach is based on the homophily assumption, where, in node classification tasks, the goal is to maximize connections between nodes of the same class while minimizing inter-class connections \cite{guo2024graphedit}. LLMs evaluate the labels between node pairs to determine whether they belong to the same class, providing suggestions on the likelihood of an edge existing between them.

LLM-GNN models utilize the semantic capabilities of LLMs to help GNNs supplement node information and understand node relationships, thereby reducing inherent noise in graph data. However, these two methods demonstrate varying performance in different scenarios: the first approach performs better in few-shot environments, while the second approach excels when textual information is abundant. Furthermore, current LLM-GNN research has a limited focus on addressing distribution shifts in graph data, which requires further exploration.

\subsection{Robustness}

Robustness refers to the ability of a system to perform consistently under various conditions. In the graph domain, robust GNNS can maintain model accuracy under perturbations such as malicious graph structure modifications. \cite{zhang2022trustworthy,zhang2023rsgnn,zhangdropedge} A robust GNN can identify and defend against malicious attacks, such as node injection and edge tampering; adapt to dynamic changes in graph structure, such as the addition or removal of nodes and edges; and maintain stable performance in the face of random noise in the data. In short, graph robustness is a key metric for measuring the consistency of GNN performance in the face of various perturbations and uncertainties.LLMs provide a new perspective and tools for enhancing the adversarial robustness of GNNs through their powerful text understanding and inferential capabilities. 


\begin{figure}
    \centering
    \includegraphics[width=1\linewidth]{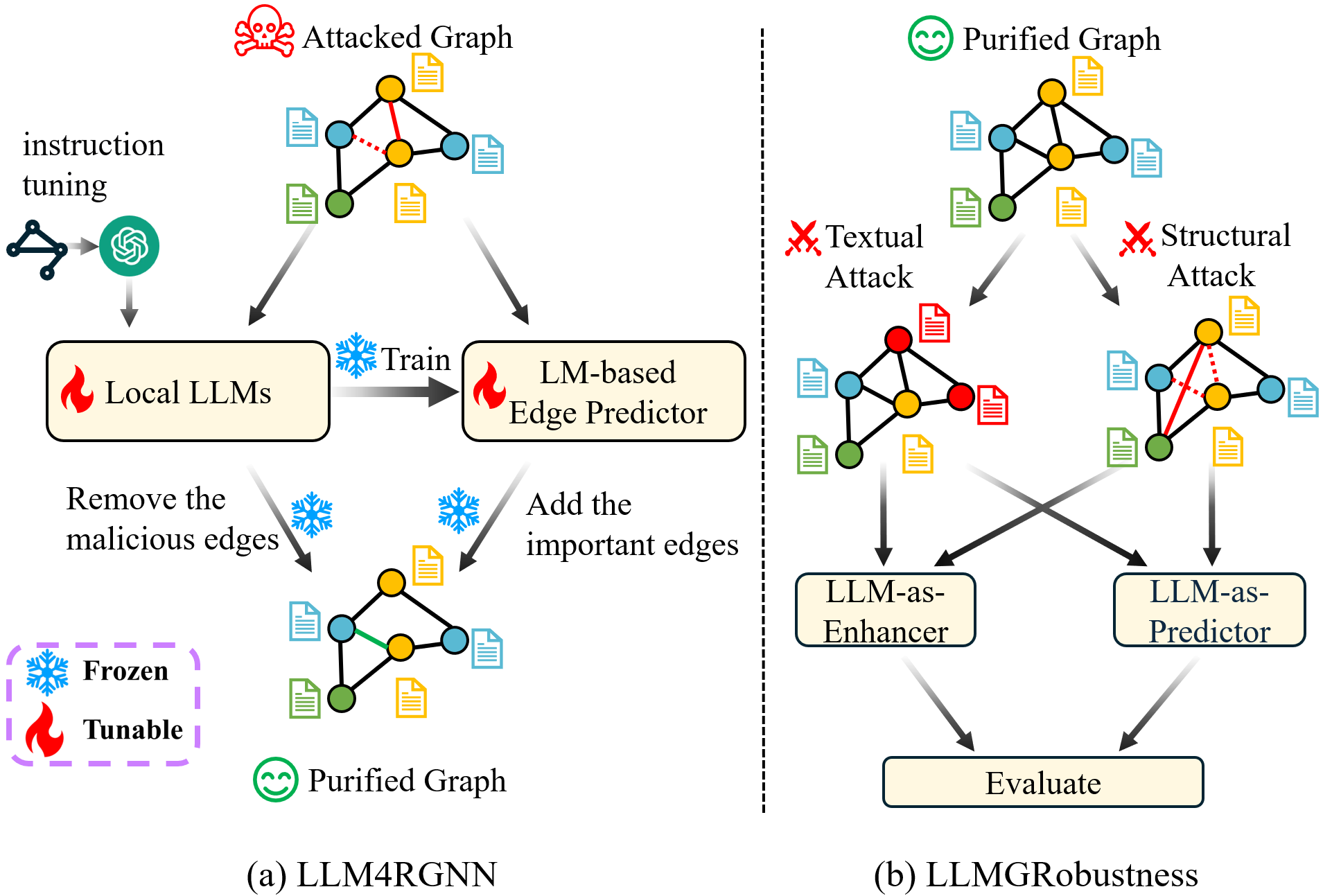}
    \caption{The illustration of LLMs integrating GNNs for improved robustness: (a) LLM4RGNN method for robustness enhancement; (b) LLMGRobustness is evaluated from two perspectives.}
    \label{fig:4}
\end{figure}

LLMs have also been used in graph-related tasks to outperform traditional GNN-based methods and produce state of the art performance, but the robustness of these LLM integrated GNN models in the face of adversarial attacks is unknown. In order to fill this research gap, Guo et al. \cite{guo2024learning} explored the robustness of LLMs in graph learning from a broad perspective, especially in the face of structural and textual perturbations.  As shown in Figure \ref{fig:4}(a), this paper uses the most popular approaches to utilize LLM for graph data, LLM-as-enhancer \cite{chen2024exploring,he2023harnessing} and LLM-as-predictor \cite{chen2024exploring,chai2023graphllm}. The experimental results of the paper indicate that LLMs, acting as enhancers or predictors, offer better robustness than shallow models. The paper also provides additional analysis to explore the potential reasons for the robustness of LLMs in graph tasks. For example, for structural attacks, the higher the quality of the features, the less dependent the model is on the structure, so the high-quality features of LLM can be more robust against structural attacks. It offers an open-source benchmark library to promote further research in this field.

The research findings suggest that although LLMs can enhance the robustness of GNNs, there is still a significant decrease in the accuracy of GNNs when faced with high-proportion topological attacks. To address this challenge proposes an edge-enhanced framework, the LLM4RGNN \cite{zhang2024can}, which takes advantage of the inferential capabilities of LLMs to identify malicious edges and recover missing important edges, thus restoring the robustness of the graph structure. As shown in Figure \ref{fig:4}(b), this framework consists of two stages: the first stage involves using instruction tuning to fine-tune a local pre-trained model to identify malicious edges; the second stage employs an LM-based edge predictor to find missing important edges and purges malicious edges based on relevance scores to restore the attacked graph structure. This framework has demonstrated consistent improvements in robustness across various GNN models, indicating the practical application potential of LLMs in enhancing GNN robustness.

Further proposals or optimizations of LLM-GNN frameworks could be explored for application in a broader range of graph data and attack scenarios. Additionally, while LLMs can provide richer node feature representations, they are less effective in dealing with graph data that lacks textual attributes, which is also an important direction for future work. 

\subsection{Privacy}
The outstanding performance of GNN models heavily depends on training with large datasets, which require collecting a vast amount of sensitive information to develop powerful GNN models capable of performing various downstream tasks, such as healthcare, bioinformatics, and banking systems \cite{wang2021gnn_finance}. However, collecting a vast amount of sensitive data from users for training GNNs raises privacy protection concerns. This issue leads to four types of attacks: membership inference attacks, inversion attacks, attribute inference attacks, and model extraction attacks.

Membership inference attacks aim to determine whether a target sample belongs to the training dataset.
Inversion attacks seek to recover the original topology of a graph or the attributes of target samples. For example, an attacker might attempt to infer the proportion of male and female users in a social network.
Model extraction attacks involve replicating the functionality or knowledge of a target model by accessing its output results. This type of attack may lead to intellectual property loss and facilitate other privacy attacks.

Traditional methods for protecting privacy in GNNs include differential privacy, federated learning, and machine unlearning:

\begin{itemize}
    \item Differential privacy introduces random noise into data or model outputs to ensure that attackers cannot ascertain whether a single sample participated in the training, thereby safeguarding data privacy.
    \item Federated learning is a distributed machine learning framework in which participants train models locally, and only model parameters are transmitted to a central server for aggregation, enabling model training without sharing raw data.
    \item Machine unlearning aims to remove the influence of specific data from a trained model. Unlike retraining the model, machine unlearning achieves efficient “deletion” through local adjustments to comply with privacy regulations such as the GDPR's “right to be forgotten.”
\end{itemize}

\begin{figure}[ht]
    \centering
    \includegraphics[width=1\linewidth]{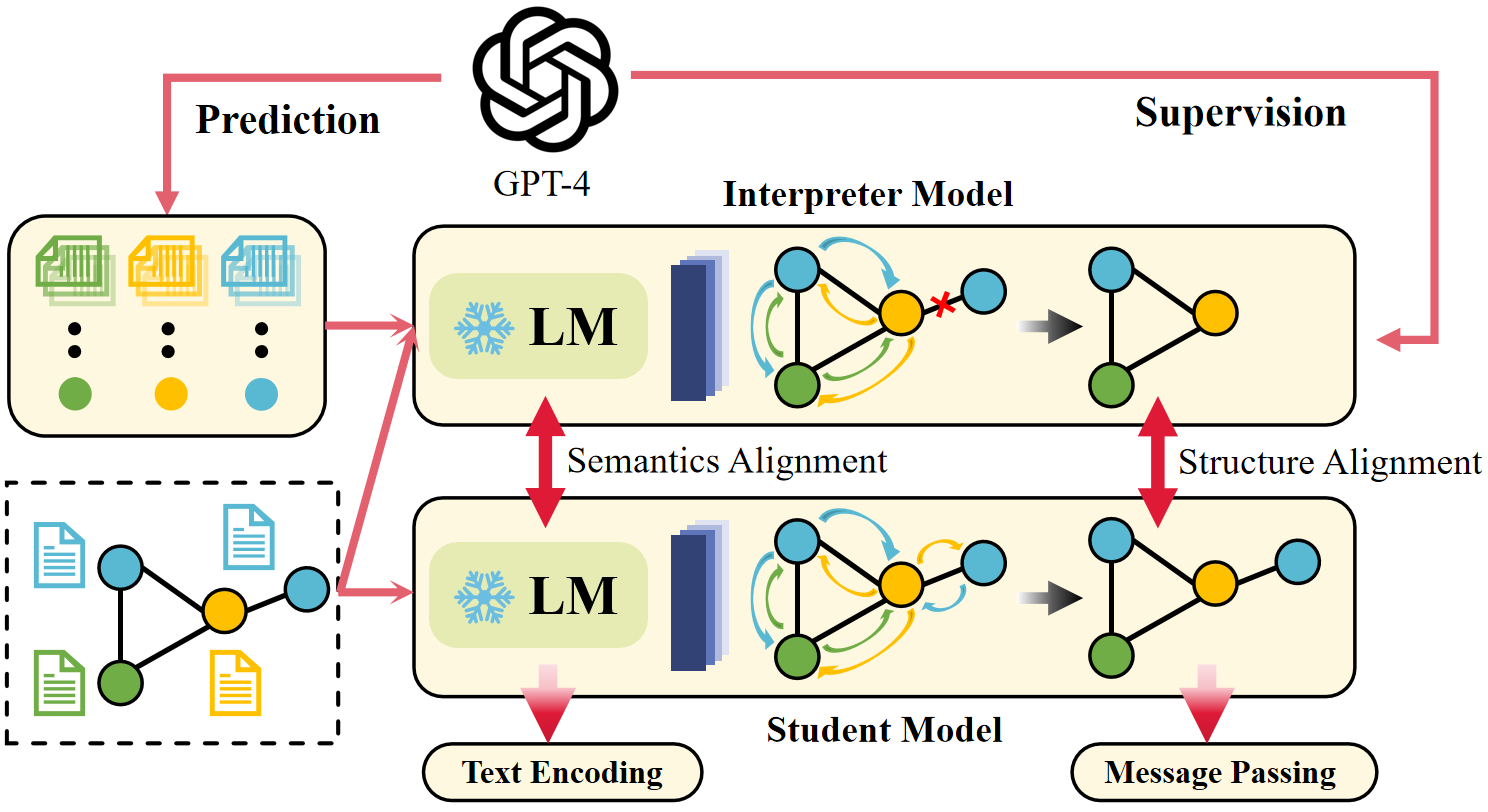}
    \caption{Knowledge Distillation Framework from LLM to Graph Model: LLMs generate pseudo labels for the LM model while simultaneously guiding the graph in identifying key nodes and key links through a supervised approach. Finally, knowledge distillation is employed to help the student model achieve superior performance}
    \label{fig:5}
\end{figure}

Research on privacy protection in LLM-GNN models is limited, and most studies focus on applying LLMs to GNN privacy attacks. Leveraging LLMs' generalization capabilities and ability to process multi-modal data, attackers fine-tune LLMs on specific datasets to create customized attack models. Additionally, in general, GNN-LLM models \cite{pan2024distilling}, users often face challenges in locally deploying LLMs due to costs and operational complexity. Instead, users typically rely on publicly available LLM APIs, which can result in privacy leaks when sensitive data is uploaded online. One existing solution is to train a GNN model using an LLM and distill its performance (in figure \ref{fig:5}) into a student model, thereby achieving LLM-optimized performance.

There is limited research on privacy protection in LLM-GNN models. In terms of protecting GNN privacy, LLMs could place greater emphasis on addressing privacy risks associated with pre-trained models. Since traditional studies on privacy attacks often focus on black-box scenarios and lack in-depth exploration of the risks of information leakage due to exposed model parameters, LLMs' strong ability to process diverse samples could be leveraged to optimize this aspect. Furthermore, personalized solutions could be developed to address the privacy protection needs of both open datasets and real-world data in the future.

\subsection{Reasoning}
Reasoning, which includes question answering, natural Language inference (NLI), and commonsense reasoning, is a fundamental skill for various NLP tasks \cite{liu2023trustworthy}. Explainability of GNNS refers to the ability to make the predictions of GNNS transparent and understandable \cite{zhang2022trustworthy}. For models that integrate GNNs with LLMs, LLMs brings reasoning capabilities to the model and also improves the explainability of the model. 

In this aspect of trustworthiness, most of the existing studies on the integration of LLMs and GNNs utilize the reasoning ability of LLMs to improve the reasoning ability and explainability of the model. Therefore, we put reasoning and explainability together.

\begin{figure}[h]
    \centering
    \includegraphics[width=1\linewidth]{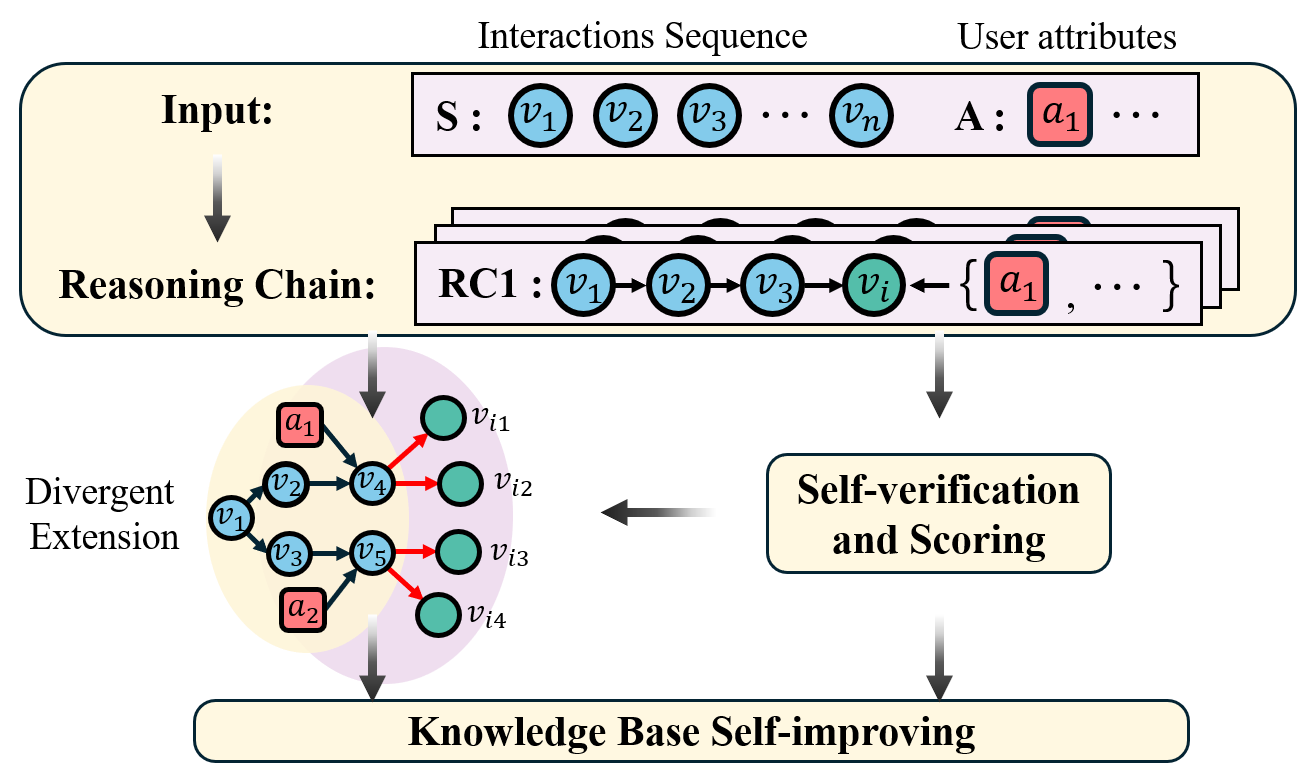}
    \caption{The illustration of LLMRG framework: LLMRG consists of four core components: chained graph reasoning, divergent expansion, self-verification and scoring, and knowledge base self-improvement.}
    \label{fig:6}
\end{figure}

LLMRG \cite{wang2023enhancing} is an innovative approach that leverages LLMs to construct personalized reasoning graphs, aiming to enhance the logical reasoning and interpretability of recommendation systems. As shown in Figure \ref{fig:6}, this framework consists of four core components: chained graph reasoning, divergent extension, self-verification and scoring, and knowledge-base self-improvement. Specifically, it initially utilizes a large language model to build a personalized reasoning graph for each user, which includes the user's interaction sequences and attribute information and infers the user's interests and behavioral motivations through causal and logical inferences. Then, the reasoning graph is encoded into a format compatible with GNNs and integrated into traditional recommendation systems, allowing recommendations to benefit from both engineered algorithms and explanatory knowledge.

Within the LLMRG framework, reasoning is the core process that enables recommendation systems to simulate and predict complex user interests and behavioral patterns. Through chained graph reasoning, the system can identify logical connections between user behaviors; through divergent extension, the system can predict potential new interests of users; self-verification and scoring ensure the rationality of the reasoning process. This reasoning-based method not only improves the accuracy of recommendations but also provides interpretability for the recommendation results, allowing the recommendation system to display the logic behind the recommendations.

\begin{figure}[h]
    \centering
    \includegraphics[width=1\linewidth]{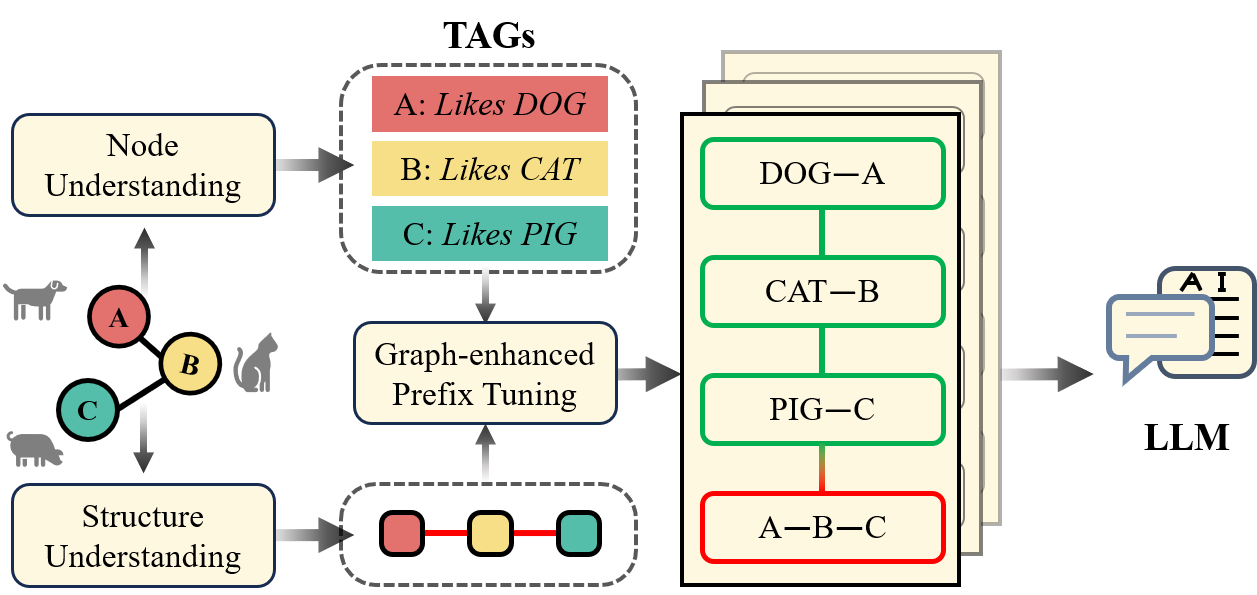}
    \caption{The illustration of GraphLLM framework:  Let the language model understand the structure of the graph directly, rather than the description of words.}
    \label{fig:7}
\end{figure}

GraphLLM \cite{chai2023graphllm} is an end-to-end method that integrates graph learning models with LLMs to enhance the reasoning capabilities of LLMs when dealing with graph data. As shown in Figure \ref{fig:7}, the framework is implemented through three main steps. First, the node understanding uses a textual transformer encoder-decoder to extract key information from the textual descriptions of nodes. Second, the structure understanding employs a graph transformer to learn the graph structure by aggregating node representations. Finally, the graph-enhanced prefix tuning transforms the graph representation into a prefix and fine-tunes it within LLMs to enhance the performance of graph reasoning tasks.

In the GraphLLM framework, reasoning is key to processing and analyzing graph data. The node understanding step allows the model to extract semantic information crucial for graph reasoning from textual descriptions; the structure understanding step enables the model to comprehend the structural relationships between nodes; and graph-enhanced prefix tuning integrates this structured information directly into LLMs, enabling the model to generate accurate responses when performing graph reasoning tasks. The reasoning capability of this method is reflected in its ability to handle complex graph structures, identify 
relationships between nodes, and make accurate predictions and decisions in various graph reasoning tasks, thereby significantly improving the performance and applicability of LLMs in graph-related tasks.

In addition, GREASELM \cite{zhang2024greaselm} adopts the LLMs multi-step inference method to construct a graph to decompose complex problems into multiple sub-problems, and LLMEP \cite{zhang2024llmexplainer} is based on the Bayesian method of LLMs to alleviate the learning bias problem in the introductory tasks of GNNs and improve the reasoning ability of the model.

In terms of explainability, VGRL \cite{ji2024vgrl} proposes a fully interpretable method, based on prompt engineering to make multiple large models collaborate with each other, and constructs a set of cyclic iteration framework. During the iteration process, the decision basis, i.e., the attributes that should be included in each label, will be continuously updated as model parameters.LLM makes decisions by input features and decision basis, and updates the decision basis of the currently involved labels according to the positive or negative prediction results, and any of the steps in this process is characterized by a natural language description, which improves the model interpretability.

Future work can focus on optimizing the construction of personalized reasoning graphs to enhance the logicality and transparency of recommender systems. It will also strive to improve the understanding of graph data by LLMs, handle more complex reasoning tasks, and optimize the efficiency and credibility of the model. These research directions will advance the progress of LLMs in the fields of graph reasoning and recommendation systems, providing more powerful technical support for practical applications.

\section{Future Research Directions}

Integrating LLMs with GNNs to improve model trustworthiness is a rapidly growing research field with a wide range of methods and applications. Given the previous review and analysis, we believe that there is still much space for improvement in this field. In this section, we outline future research directions and highlight the great potential of utilizing the integration of LLMs and GNNs to boost other aspects of model trustworthiness.

\textbf{Privacy Protection and Robustness Enhancement.} Current LLM-GNN models face conflicts between traditional privacy protection methods, such as differential privacy and federated learning, and the characteristics of LLMs. In the future, it is necessary to explore collaborative training frameworks that enhance privacy, for example, by reducing the exposure of sensitive data through localized model distillation, or by leveraging the reasoning capabilities of LLMs to proactively identify adversarial attacks. Additionally, to address the decline in robustness caused by high-proportion topological attacks, further optimization of LLM-driven structural repair methods is needed. This can be achieved by combining dynamic edge enhancement and semantic verification to improve the stability of models under noise and adversarial perturbations. Moreover, there is a need to address the robustness challenges of non-textual graph data, such as molecular networks, and explore cross-modal alignment and structure-driven LLM adaptation strategies. It is worth mentioning that studies have pointed out that it is necessary to consider complex cross-aspect relationships when building trustworthy GNN systems \cite {zhang2022trustworthy}. Therefore, when constructing a credible LLM-GNN model in the future, it is necessary to balance the conflicts between the credibility dimensions. For example, strict privacy protection may reduce model robustness, and future research needs to consider credible cross-aspect relationships.

\textbf{Trusted LLM-GNN with Low Text Dependency.} Many existing works on integrating LLMs with GNNs to enhance trustworthiness focus on textual attribute graphs, or describe the nodes and relationships between nodes clearly with text, which reflects the high dependence on text in existing works. The increased workload caused by heavy text dependence, along with the need for broader application scenarios, highlights the need to develop trusted LLM-GNN models with reduced text dependence. An important direction for future research is to explore the integration of image features with graph structural data using LLMs for multimodal learning to improve model generalization. Another promising direction is to explore methods for handling text-free graphs, such as molecular structures, by leveraging the prior knowledge of LLMs. Trustworthy LLM-GNN models with reduced text dependence have broad applications. For example, gene interaction analysis and protein structure prediction in the biomedical field can be modeled as graph-based tasks with minimal textual information. Applying LLM-GNN models to these areas could lead to breakthroughs in biomedical research.

\textbf{Opportunities for the Rapid Development of LLMs.} LLMs have developed rapidly in recent years and have produced tremendous changes, which have a great impact on academic research and people's production and life. The continued advancement of LLMs is expected to significantly enhance research on the trustworthiness of LLM-GNN models. With the advent of multimodal LLMs, such as GPT-4V \cite{yang2023dawn}, which facilitates the processing of graph data with image nodes. This suggests that future developments in LLMs will enable credible LLM-GNN models to support a wider range of applications while improving efficiency.


\textbf{Fairness.} As an important subfield of trustworthy, fairness-aware GNN aims to reduce bias to ensure fairness in the predictions of different groups. These groups are divided based on a multicomponent sensitive attribute. 
The fundamental assumptions and design of GNNs often lead to models achieving superior accuracy in certain groups. No work has applied LLMs to fairness-aware GNNs so far, so we propose some feasible ideas. Fairness-aware GNN can be designed to achieve fairness through different objectives. (1)\textit{Graph-level Fairness}: connections between different sensitive attribute groups are balanced. (2)\textit{Neighborhood-level Fairness}: node's neighborhood has a balanced distribution of sensitive attributes. (3)\textit{Embedding-level Fairness}: node representations generated by GNN leak no sensitive attribute information. (4)\textit{Prediction-level Fairness}: the final predictions are fair to different groups. The first two directions typically address bias at the graph or neighborhood level by modifying the graph structure before or during preprocessing, such as deleting highly homogeneous edges \cite{spinelli2021fairdrop}, increasing the weight of heterogeneous neighbors \cite{chen2022graph,yang2024fairsin}, or resampling to balance the distribution of sensitive attributes. However, these strategies for fairness enhancement in pure graph structures may not suit real-world applications. A fair model should go beyond simple data processing and focus on Text-Attributed Graphs, Text-Paired Graphs, or Multimodal Learning. LLMs can be used as fair data encoders to remove sensitive attributes from the paired text of nodes or to describe node text without leaking sensitive information. The latter two directions focus on designing equity-aware loss functions \cite{bose2019compositional} or disentangling sensitive attributes from embeddings \cite{zhu2024fair}. However, the prior knowledge of the determined sensitive attribute has too strong a hypothesis, which makes it hard to judge whether the final embedding space covers other sensitive attributes that have not been paid attention to, and also leads to worsened interpretability. Promising solutions are to use LLMs as fairness aligners to supervise GNN-generated embeddings or as evaluators to assess model fairness, leveraging their textual reasoning capabilities to provide interpretable insights.

\section{Conclusion}
This paper systematically reviews recent developments in the field of trustworthy graph neural networks enhanced by large language models (LLM-GNNs), with a particular focus on the application of LLM technology in trustworthy GNN scenarios, such as reasoning, privacy, robustness, and reliability, as well as the key challenges in these areas. The semantic capabilities of LLMs significantly enhance trustworthy GNNs’ ability to handle erroneous information. However, their impact on optimizing traditional privacy-preserving methods, such as differential privacy and federated learning, remains limited, indicating the need for further in-depth exploration of these techniques. Looking ahead, as LLM technology advances, challenges in data processing and model trustworthiness are expected to be further mitigated through more accurate and reliable responses from LLMs. This review offers researchers a comprehensive perspective on trustworthy LLM-GNNs, fostering both theoretical and practical advancements in the field.


\bibliographystyle{named}
\bibliography{ijcai25}

\end{document}